\documentclass[10pt,twocolumn,letterpaper]{article}

\usepackage[pagenumbers]{cvpr} 

\usepackage{graphicx}
\usepackage{amsmath}
\usepackage{amssymb}
\usepackage{booktabs}

\usepackage{amsmath}
\usepackage{pifont}
\newcommand{\cmark}{\text{\ding{51}}}
\newcommand{\xmark}{\text{\ding{55}}}

\usepackage[table,x11names]{xcolor}

\usepackage[pagebackref,breaklinks,colorlinks]{hyperref}

\usepackage[capitalize]{cleveref}
\crefname{section}{Sec.}{Secs.}
\Crefname{section}{Section}{Sections}
\Crefname{table}{Table}{Tables}
\crefname{table}{Tab.}{Tabs.}

\def\cvprPaperID{*****} 
\def\confName{CVPR}
\def\confYear{2023}

\begin{document}

\title{EPIC-KITCHENS-100 Unsupervised Domain Adaptation
Challenge: \\Mixed Sequences Prediction}



\author{Amirshayan Nasirimajd, Simone Alberto Peirone, Chiara Plizzari, Barbara Caputo\\
Politecnico di Torino\\
{\tt\footnotesize amirshayan.nasirimajd@studenti.polito.it, \{simone.peirone, chiara.plizzari, barbara.caputo\}@polito.it} \\
}

\maketitle

\begin{abstract}
This report presents the technical details of our approach for the EPIC-Kitchens-100 Unsupervised Domain Adaptation (UDA) Challenge in Action Recognition.
Our approach is based on the idea that the order in which actions are performed is similar between the source and target domains. Based on this, we generate a modified sequence by randomly combining actions from the source and target domains. As only unlabelled target data are available under the UDA setting, we use a standard pseudo-labeling strategy for extracting action labels for the target. We then ask the network to predict the resulting action sequence. This allows to integrate information from both domains during training and to achieve better transfer results on target. 
Additionally, to better incorporate sequence information, we use a language model to filter unlikely sequences.
Lastly, we employed a co-occurrence matrix to eliminate unseen combinations of verbs and nouns.
Our submission, labeled as `sshayan', can be found on the leaderboard, where it currently holds the 2nd position for `verb' and the 4th position for both `noun' and 'action'.
\end{abstract}


\label{sec:intro}
\section{Introduction}
Egocentric action recognition, aiming to understand human activities from a first-person view, has gained significant attention with the rising popularity of wearable devices. 
Those have provided access to a vast amount of data, creating opportunities for applications in augmented reality, robotics, and human-computer interaction.

However, achieving high accuracy on new data 
becomes challenging due to the so called \textit{domain shift}, which refers to the difference in terms of distribution between the training (source) and test set (target), caused by changes in visual appearance or contexts, leading to a significant performance gap. To address this challenge, the Unsupervised Domain Adaptation (UDA) setting has been proposed, where an unlabeled set of samples from the test distribution is available during training for adaptation. 

\begin{figure}[t]
  \centering
  \includegraphics[width=\linewidth]{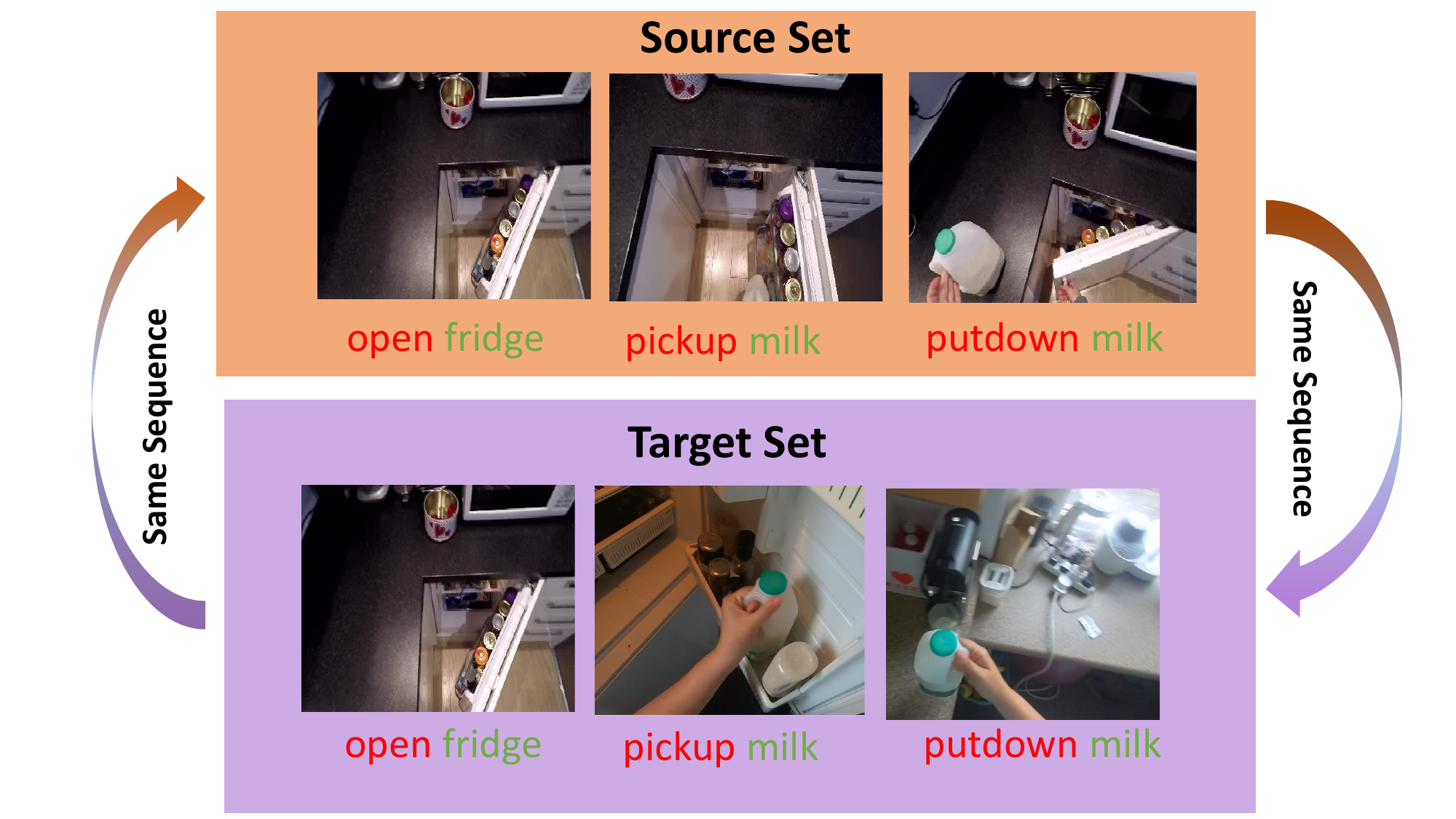}
   \caption{Example of action sequence similarities across two different kitchens. When ``\textit{picking something out of the fridge}", the sequence ``open", ``pick-up", and ``put-down" repeats across the two domains. }
   \label{fig:teaser}
\end{figure}

Human activities tend to follow patterns as actions are strongly influenced by the temporal context in which they are performed. For example, regardless of the person and the environment, \textit{putting something in the fridge} always involves the same steps, as shown in Figure~\ref{fig:teaser}.
Based on this simple observation, we propose to use sequential prediction and the inherent relationships between action sequences to mitigate the negative effects of domain shift in EAR.

\begin{figure*}[t]
  \centering
  \includegraphics[trim={0 6cm 0 7cm}, width=\linewidth]{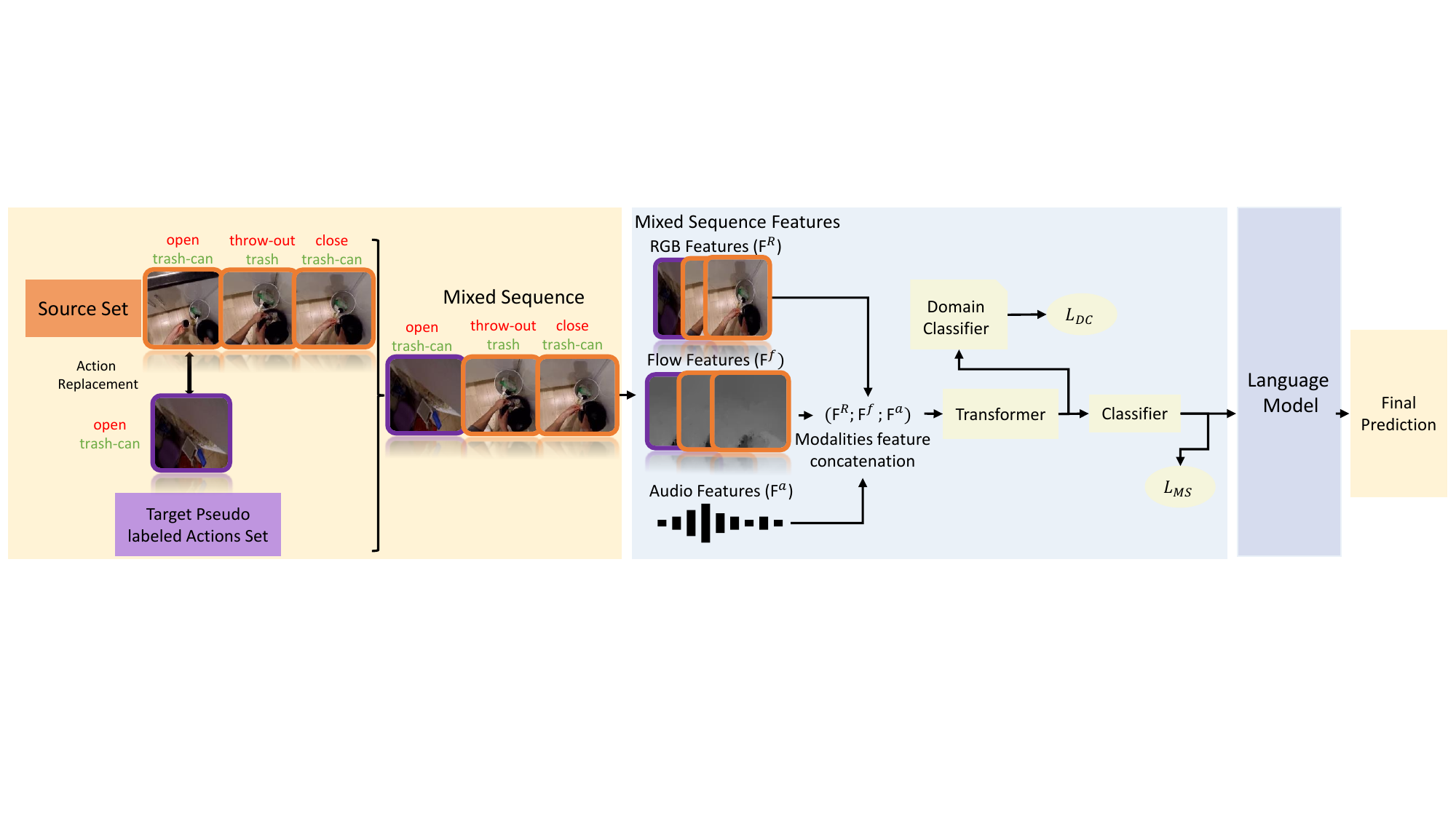}
   \caption{Overall architecture of the proposed framework. First, we randomly choose an action among the $n$ actions of the sequence from the source domain. Next, we use pseudo-labels on the target to find a substitute for the selected action sharing the same verb and noun labels. 
   RGB, optical flow and audio features of the samples in the sequence are fed to a transformer to enable information sharing between neighbouring actions.
   A domain classifier encourages the network to produce domain agnostic features. 
   Finally, we use the language model to refine the sequence predictions.
   }
   \label{fig:method}
\end{figure*}



Our approach consists in replacing random action clips in a source domain sequence with action clips from the target domain that represent the same action. By training the model to predict the sequence of actions in the modified sequences, we encourage the model to learn common sequence patterns that exist across different domains, aiming at mitigating the influence of domain shift. 
Additionally, by taking inspiration from~\cite{kazakos_temporal2021}, we integrate a language model into our framework to better incorporate the context derived from the surrounding actions. 
Finally, we compute a co-occurrence matrix on the action of the source domain to filter out improbable predictions of verb and noun pairs, thus further refining the prediction process, as done in~\cite{cheng2022team}.


We evaluate our approach on the EPIC-Kitchens Unsupervised Domain Adaptation challenge~\cite{Damen2022RESCALING}, showcasing the effectiveness of exploiting sequential information in addressing the domain gap. 

\section{Our Approach}

In this section, we describe our method in details. 
We introduce our approach to model temporal dependencies between actions and to adapt action sequences across different domains. 
Finally, we refine the predictions of our model using a language model and we reduce the likelihood of actions not seen during training.
Figure~\ref{fig:method} presents the different components of our method.

\subsection{Mixed Sequence Generation}
The main objective of our method is to augment action sequences from the source domain with samples from the target sharing the same verb and noun labels.
As labels are not available in the target domain, we assign \textit{pseudo-labels} to the target data, based on a confidence threshold $\lambda$. 
For each target sample, we define the confidence of the model predictions as $(p_v + p_n) / 2$, where $p_v$ and $p_n$ represent the probabilities of the top verb and noun predictions respectively. 
Samples whose prediction confidence falls below the threshold are not considered in the mixing process.

Given a sequence consisting of $w$ actions, our goal is to predict the central action $i$ within the temporal window $(i - w/2 < i < i + w/2)$, while exploiting the context provided by both preceding and subsequent actions.

To incorporate target domain information, we introduce a domain adaptation approach for replacing action clips from the source domain with action clips from target distribution representing the same action. More specifically, for each sequence of actions $\mathcal{S}_i=\{s_{-w/2}, \dots, s_0, \dots s_{w/2}\}$ from the source domain, we randomly select one or more item ($s_i$) - excluding the central action - and replace them with an equivalent $t_i$ sampled from the target domain, with the constraint that $t_i$ pseudo-labels match verb and noun labels of $s_i$. The model is trained to predict the verb and noun labels of the central action of the mixed sequence $\mathcal{\Tilde{S}}_i=\{s_{-w/2}, ..., t_i, ..., s_{w/2}\}$.

\subsection{Sequence Predictor }
To integrate sequences in the action recognition process, we take inspiration from~\cite{kazakos_temporal2021} to model temporal relations within the actions window using a transformer model.

\textbf{Positional Embedding}
In the first step, we concatenate input features of different modalities and project the output to a lower dimension $D$. 
Then, we apply learnt positional embedding to tag the position of each action in the temporal window. Finally, we also append to the sequence two separate tokens for verb and noun prediction of central action. The resulting sequence is indicated by $X_e$. 

\textbf{Transformers Encoding}
Once temporal consistency is encoded in positional embeddings, the resulting sequence is fed to a transformer encoder $f(.)$. The attention layers in the transformer allow actions inside the sequence to attend to all the surrounding actions, possibly exchanging relevant clues for the classification of the central action. This operation can be expressed as $Z = f(X_e)$, where $Z$ represents the output of the transformer module, and $X_e$ is the output of positional embedding. 


\textbf{Action Classifier}
Finally, samples in the sequence are classified using two heads, to predict the verb and noun labels separately. 
As in~\cite{kazakos_temporal2021}, we define two classification losses. 
The primary loss function is defined as the cross entropy of the central action of the sequence, computed separately for verb and noun predictions.
Then, to further encourage the network to model temporal relations between all neighbouring actions in the sequence, we also ask the model to predict the labels for all the actions in the sequence through standard cross entropy loss, which we refer to as the mixed sequence prediction loss ($\mathcal{L}_{MS}$).

To encourage the network to learn domain agnostic representations, we also include a domain adversarial module with Gradient Reversal Layer~\cite{Ganin2015} to produce a domain classification loss $\mathcal{L}_{DC}$ for all the actions in the sequence. 
The resulting loss is thus determined by the summation of the domain discriminator loss $\mathcal{L}_{DC}$ and the sequence prediction loss $\mathcal{L}_{MS}$.

\subsection{Language Model}
Inspired by MTCN~\cite{kazakos_temporal2021}, we introduce a Language Model (LM) to filter out unlikely action sequences. 
We adopt the Masked Language Model (MLM) approach as in~\cite{kazakos_temporal2021}, in which the model is trained to predict the masked actions within a sequence from the source data, thus encouraging the model to derive high-level dependencies between the actions that make up the sequence. 
At inference time, to integrate the LM into our framework, we generate all possible sequences of size $w$ using the top-$k$ predictions of the mixed sequence model for each action and select the most probable sequence according to the LM. 
The final prediction of the model is a linear combination of the predictions from the mixed sequence $y_{MS}$ and the output of the language model for the most probable sequence $y_{LM}$:
\begin{equation}
y_{final} = (1 - \beta)y_{MS} + \beta y_{LM}.
\label{eq:LM}
\end{equation}

\subsection{Co-Occurrence}
To further refine the model predictions, we compute a co-occurrence matrix $\mathcal{M}_{CO}$ of verbs and nouns labels in the source domain, as introduced in~\cite{cheng2022team}.
Each entry $\mathcal{M}_{CO_{i,j}}$ stores the occurrences of verb class $i$ and noun class $j$ together in the source data.
At test time, the probability of predicting verb and noun pairs which are not present in the source data reduces by a factor of $0.01$, i.e. we use the simple assumption that actions not present in the source data are unlikely to be found in the target data.

\section{Experiments}

\subsection{Implementation Details}

\begin{table*}
    \centering
    \small
    \begin{tabular}{ccccc|ccc}
    \toprule
        Sequence len $>$ 1 & $\mathcal{L}_{MS}$ & $\mathcal{L}_{DC}$ & LM & $\mathcal{M}_{CO}$ & Verb  & Noun  & Action \\
        \midrule
         \xmark & \xmark & \xmark & \xmark & \xmark &  51.17 & 28.88 & 20.93 \\ 
         \cmark & \xmark & \xmark & \xmark & \xmark &  51.27 & 30.48 & 21.59 \\ 
         \cmark & \cmark & \xmark & \xmark & \xmark &  53.22 & 32.68 & 23.40 \\ 
         \cmark & \cmark & \cmark & \xmark & \xmark &  54.30 & 33.31 & 24.58 \\ 
         \cmark & \cmark & \cmark & \cmark & \xmark &  \textbf{55.22} & 34.07 & 24.99 \\ 
         \cmark & \cmark & \cmark & \cmark & \cmark &  54.95 & \textbf{34.15} & \textbf{25.46} \\ 
         \bottomrule
    \end{tabular}
    \caption{Ablation study of the different components of our architecture on TBN features. The table shows the Top-1 accuracies starting from the baseline, e.g. each action sample is classified individually, and adding sequence predictions with a temporal window of 5 actions, mixing with the target data ($\mathcal{L}_{MS}$), adversarial alignment using a domain classifier ($\mathcal{L}_{DC}$), predictions refinement using a language model (LM) and pruning of the unlikely verb and noun pairs using a Matrix of Co-Occurrences ($\mathcal{M}_{CO}$). 
    }
    \vspace{-10pt}
    \label{tab:tbn}
\end{table*}

\textbf{Sequence Predictor Architecture. }
To train the mixed sequence model, we use the pre-extracted Temporal Binding Network (TBN)~\cite{kazakos2019TBN} features from the Unsupervised Domain Adaptation (UDA) splits of the EPIC-Kitchens dataset.
These features were available for three modalities: RGB, Flow, and Audio. Each modality comprises 25 clips with features size 1024 and we use TRN~\cite{zhou2018temporal} to temporally aggregate the sample clips.

The model is trained for 100 epochs using a SGD optimizer, with learning rate 0.005. Pseudo-labels for the target data are generated with a confidence threshold of $\lambda$ = 0.75. During each epoch, we mix actions between the source and target domains. Specifically, we replace one action from the source domain with a corresponding action from the target domain using a temporal window of length 5.

\textbf{Language Model. }
The language model is trained for an additional 100 epochs using the Adam optimizer, with a loss value set to 0.001, and fine-tuned on the source training set labels. The training process incorporates a temporal window ($w$) of size 5. The weight coefficient ($\beta$) that combines the predictions of the language model and sequence predictor is set to 0.25. To score sequences, the top-5 actions predicted by the model are taken into account.

\subsection{Results}

Table~\ref{tab:tbn} presents an ablation study of our method, using three modalities and TBN features. The sequence attention mechanism from MTCN~\cite{kazakos_temporal2021} enhances the accuracy compared to the baseline, highlighting the significance of temporal context reasoning even in the presence of domain shift. Furthermore, the inclusion of target and source mixing further boosts the accuracy. 
Additionally, the introduction of a domain classifier aids in better integration of target information into our model, resulting in further improvement. 
Subsequently, we apply to filter improbable predictions using the language model, leading to enhanced action accuracy, and demonstrate the role of temporal context in alleviating the domain gap. Lastly, the inclusion of the co-occurrence matrix yields a final improvement to the model's performance, especially on the \textit{action} metric as it weakens the probability of the unlikely verb and noun pairs.

\subsection{Ablations}
\textbf{Sequence length. }
Table~\ref{tab:sequence_length} shows the effect of the number of actions $w$ in the temporal window on verb and noun classification accuracy, without mixing actions between source and target. Best results on action category are obtained by using a sequence of length $w=5$. 
\begin{table}
    \centering
    \small
    \begin{tabular}{c|ccc}
        \toprule
        Sequence length & Verb  & Noun  & Action \\
        \midrule
        1 & 51.17 & 28.88 & 20.93 \\
        \hline
         3 &  50.97 & 30.23 & 21.18 \\ 
         \hline
         5 &  51.27 & \textbf{30.48} & \textbf{21.59} \\ 
         \hline
         
         9 & \textbf{51.31} & 30.34 & 21.41\\
         \bottomrule
    \end{tabular}
    \caption{Top-1 accuracy using different numbers of actions in the temporal window $w$ on the EPIC-Kitchens-100 validation set.}
    \label{tab:sequence_length}
\end{table}

\textbf{Number of replacements. } 
Table~\ref{tab:mixing_level} shows the effect of replacing one or more actions in the sequence with the target before the sequence is sent to the transformer. Improvements in accuracy on all metrics are observed when at least one sample is substituted, while the substitution of more samples only improves on individual categories.
\begin{table}
    \centering
    \small
    \begin{tabular}{c|ccc}
        \toprule
        \# Replacements & Verb  & Noun  & Action \\
        \midrule
         0 &  51.27 & 30.48 & 21.59 \\ 
        \hline
         1 &  53.22 & 32.68 & 23.40 \\ 
        \hline
         2 & 53.13 & 32.07 & \textbf{23.73}\\
        \hline
         3 & \textbf{53.64} & \textbf{32.70} & {23.59}\\
        \bottomrule
    \end{tabular}
    \caption{Top-1 accuracy using a different number of target replacements results within a sequence of $w=5$ actions. Results reported on the EPIC-Kitchens-100 validation set.}
    \label{tab:mixing_level}
\end{table}

\subsection{Model Ensemble}
For the final submission, we ensemble our technique with different backbones, using SlowFast~\cite{fan2020pyslowfast} and Temporal Shift Module~\cite{lin2019tsm} trained on EPIC-Kitchens-55 with ResNet50.
The performances of the individual backbones and the ensemble are presented in Table~\ref{tab:ensemble}.
\begin{table}
    \centering
    \small
    \begin{tabular}{c|ccc}
        \toprule
        Backbone & Verb  & Noun  & Action \\
        \midrule
         TBN~\cite{kazakos2019TBN} &  54.30 & 33.31 & 24.58 \\ 
         \hline
         TSM~\cite{lin2019tsm} &  54.13 & 33.25 & 24.53 \\ 
         \hline
         SlowFast~\cite{fan2020pyslowfast} & 54.44  & 30.74 & 23.38  \\ 
         \hline
         Ensemble(E) & 56.97 & 35.64 & 26.50\\
         \hline
         E + LM & \textbf{57.46} & \textbf{36.44} & 27.25\\
         \hline
         E + LM + $\mathcal{M}_{CO}$ & 57.24 & 36.42 & \textbf{27.63}\\
         \bottomrule
    \end{tabular}
    \caption{Top-1 accuracy using our domain adaptation method with different backbones on the EPIC-Kitchens-100 validation set. LM: Language Model. $\mathcal{M}_{CO}$: Matrix of Co-Occurrences.}
    \vspace{-10pt}
    \label{tab:ensemble}
\end{table}
Our approach is visible on the official leaderboard and shown in Table~\ref{tab:leaderboard}, along with the top five teams' performance.
\begin{table}
    \centering
    \small
    \begin{tabular}{c|c|ccc}
        \toprule
        Rank & Method & Verb  & Noun  & Action \\
        \midrule
         1&Ns-LLM &  \textbf{58.22} & 40.33 & \textbf{30.14} \\ 
         \hline
         2&VI-I2R &  57.89 & 40.07 & 30.12 \\ 
         \hline
         3&Audio-Adaptive-CVPR2022 &  52.95 & \textbf{42.26} & 28.06 \\ 
        \hline
        \rowcolor{lightgray}
        \textbf{4} & \textbf{sshayan} &  58.11 & 35.89 & 27.72 \\ 
        \hline
        5&plnet & 55.51 & 35.86 & 25.25\\
        \bottomrule
    \end{tabular}
    \caption{Top-1 accuracy on the official leaderboard of the UDA challenge EPIC-Kitchens-100. Our submission is highligthed.}
    \vspace{-8pt}
    \label{tab:leaderboard}
\end{table}

\section{Conclusion}
In conclusion, this report presents our sequential approach to the EPIC-Kitchens-100 unsupervised domain adaptation challenge. We propose a mixed sequence strategy to improve the transferability of our model across domains. Furthermore, we leverage a language model and co-occurrence matrix to integrate contextual information by filtering out improbable combinations of verbs and nouns. Through these techniques, our model achieves significant performance improvements.
{\small
\bibliographystyle{ieee_fullname}
\bibliography{egbib}
}

\end{document}